%% file: main.tex
\documentclass[runningheads]{llncs}
\usepackage{eccv}
\usepackage{eccvabbrv}
\usepackage{graphicx}
\usepackage{booktabs}
\usepackage{multirow}
\usepackage{pifont}
\usepackage[accsupp]{axessibility}
\usepackage[pagebackref,breaklinks,colorlinks,citecolor=eccvblue]{hyperref}
\usepackage{hyperref}
\usepackage{orcidlink}

\begin{document}

\title{WAFT-Stereo: Warping-Alone Field Transforms for Stereo Matching} 

\titlerunning{WAFT-Stereo: Warping-Alone Field Transforms for Stereo Matching}

\author{Yihan Wang \and Jia Deng}

\authorrunning{Wang and Deng}
\institute{Department of Computer Science, Princeton University
\\ 
\email{\{yw7685, jiadeng\}@princeton.edu}}

\maketitle

\input{macros}

\begin{abstract}
  We introduce WAFT-Stereo, a simple and effective warping-based method for stereo matching. WAFT-Stereo demonstrates that cost volumes, a common design used in many leading methods, are not necessary for strong performance and can be replaced by warping with improved efficiency. WAFT-Stereo ranks first on ETH3D (BP-0.5), Middlebury (RMSE), and KITTI (all metrics), reducing the zero-shot error by 81\% on ETH3D, while being $1.8-6.7\times$ faster than competitive methods. Code and model weights are available at \href{https://github.com/princeton-vl/WAFT-Stereo}{https://github.com/princeton-vl/WAFT-Stereo}.
\end{abstract}

\begin{figure}[t]
    \centering
    \includegraphics[width=1.0\linewidth]{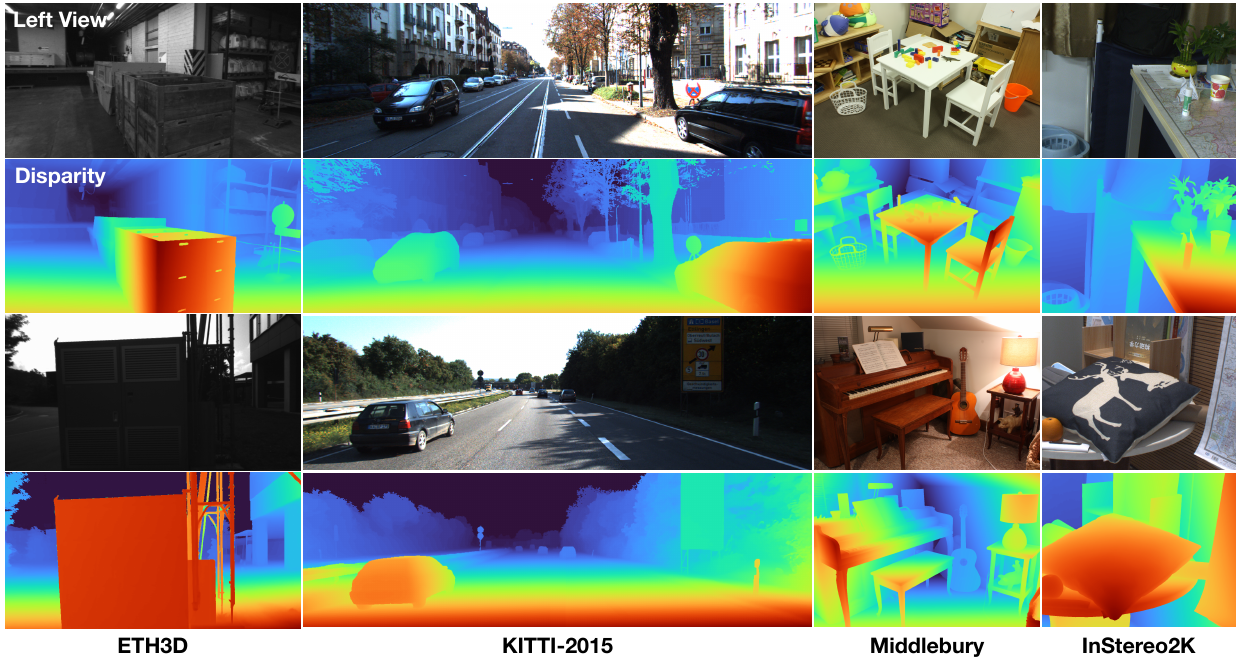}
    \caption{WAFT-Stereo achieves strong sim-to-real generalization~\cite{eth3d, middlebury, menze2015object, bao2020instereo2k}.}
    \label{fig:sim2real-vis}
\end{figure}

\section{Introduction}
\label{sec:intro}

Stereo matching is the problem of estimating the horizontal motion of each pixel given two rectified images from two calibrated cameras. The horizontal motion, or disparity, can be directly converted to depth. Stereo matching has many direct applications, including autonomous driving~\cite{geiger2012we, menze2015object} and augmented reality~\cite{engel2023project}.

Most leading stereo matching methods~\cite{wen2025foundationstereo, lipson2021raft, igev, cheng2025monster++, min2025s2m2, guan2025bridgedepth, jiang2025defom} rely on cost volumes~\cite{sun2018pwc, zbontar2015computing}, which are constructed by pairwise comparisons across the two input views.  While widely adopted, cost volumes incur substantial memory overhead that scales linearly with the disparity range (full cost volume) or look-up radius (partial cost volume), and are therefore typically constructed and processed at low resolution (e.g., 1/4 of the original resolution). However, low resolution can hurt accuracy, especially for highly detailed image regions. 

Stereo matching can be understood as a special case of optical flow that is restricted to the horizontal scan line. Recently, WAFT~\cite{wang2025waft}, a new state-of-art optical flow estimator, demonstrates that cost volumes can be replaced by high-resolution feature-space warping, leading to a substantially simpler design and improved accuracy and efficiency. This raises a natural question: \emph{can cost volumes be replaced by warping also for stereo matching?}

In this work we answer this question in the affirmative. We introduce WAFT-Stereo, a warping-based stereo matching method that achieves a new state of the art. WAFT-Stereo demonstrates that strong performance does not require cost-volume-specific designs. Instead, high-resolution warping paired with iterative updates~\cite{teed2020raft, lipson2021raft,wang2025waft} are sufficient.

WAFT-stereo is based on WAFT, but involves non-trivial modifications. WAFT can be directly applied to stereo matching---just changing the prediction to 1D flow---but this trivial adaptation does not work well in practice.  WAFT uses recurrent iterative updates which regress updates to flow vectors. This works well for small displacements between adjacent frames of continuous video, but can struggle to converge for very large displacements (hundreds of pixels) typical for high resolution stereo pairs. 

WAFT-Stereo addresses the issue of large displacements by introducing a classification module before recurrent updates. The classification module predicts disparity discretized into a predefined set of bins. The classification module uses the same architecture as the recurrent updater. This is a small change to the prediction head and extremely simple to implement, yet it significantly improves accuracy at negligible additional cost.

In addition, WAFT-Stereo introduces several architectural improvements over WAFT. WAFT includes a small U-Net that serves an lightweight adaption layer for a pre-trained input encoder. We remove this U-Net by instead fine-tune the pre-trained encoder with LoRA~\cite{hu2022lora}, which reduces latency. Also, we replace the high-resolution skip connection with several ResNet blocks in the recurrent update module, which significantly improves accuracy.

WAFT-Stereo achieves state of the art accuracy on standard benchmarks. This was achieved using the same DAv2-L backbone~\cite{yang2024depth} used by leading prior stereo matching approaches (e.g., FoundationStereo~\cite{wen2025foundationstereo}, DEFOM-Stereo~\cite{jiang2025defom}, and Monster++~\cite{cheng2025monster++}). WAFT-Stereo reduces BP-0.5 by 6\% on ETH3D~\cite{eth3d}, BP-2 by 13\% on KITTI-2012~\cite{geiger2012we}, D1 by 6\% on KITTI-2015~\cite{menze2015object}, and RMSE by 6\% on Middlebury~\cite{middlebury}.

WAFT-Stereo is highly efficient. It processes 540p stereo pairs at 10 FPS on an NVIDIA L40 GPU, representing a $1.8$--$6.7\times$ speedup over leading methods (e.g., $1.8\times$ over S2M2-XL~\cite{min2025s2m2} and $6.7\times$ over FoundationStereo~\cite{wen2025foundationstereo}). With a smaller backbone, DAv2-S~\cite{yang2024depth}, WAFT-Stereo reaches 21 FPS on 540p inputs while maintaining accuracy comparable to prior state-of-the-art methods.

WAFT-Stereo also exhibits strong generalization, when evaluated in the zero-shot setting. Trained exclusively on synthetic data, it achieves the best BP-0.5 on the ETH3D benchmark among all existing submissions, corresponding to an 81\% error reduction over the strongest established zero-shot baseline~\cite{wen2025foundationstereo}. On KITTI-2015, WAFT-Stereo reduces D1 by 9\% relative to leading zero-shot methods~\cite{wen2025foundationstereo, min2025s2m2}.

Our main contributions are two-fold: (1) we show that cost volumes are not necessary for strong performance in stereo matching, and (2) we introduce WAFT-Stereo, a fully warping-based architecture with state-of-the-art accuracy, high efficiency, and strong generalization.

\section{Related Work}
\label{sec:relwork}

\begin{figure}[t]
    \centering
    \includegraphics[width=1.0\linewidth]{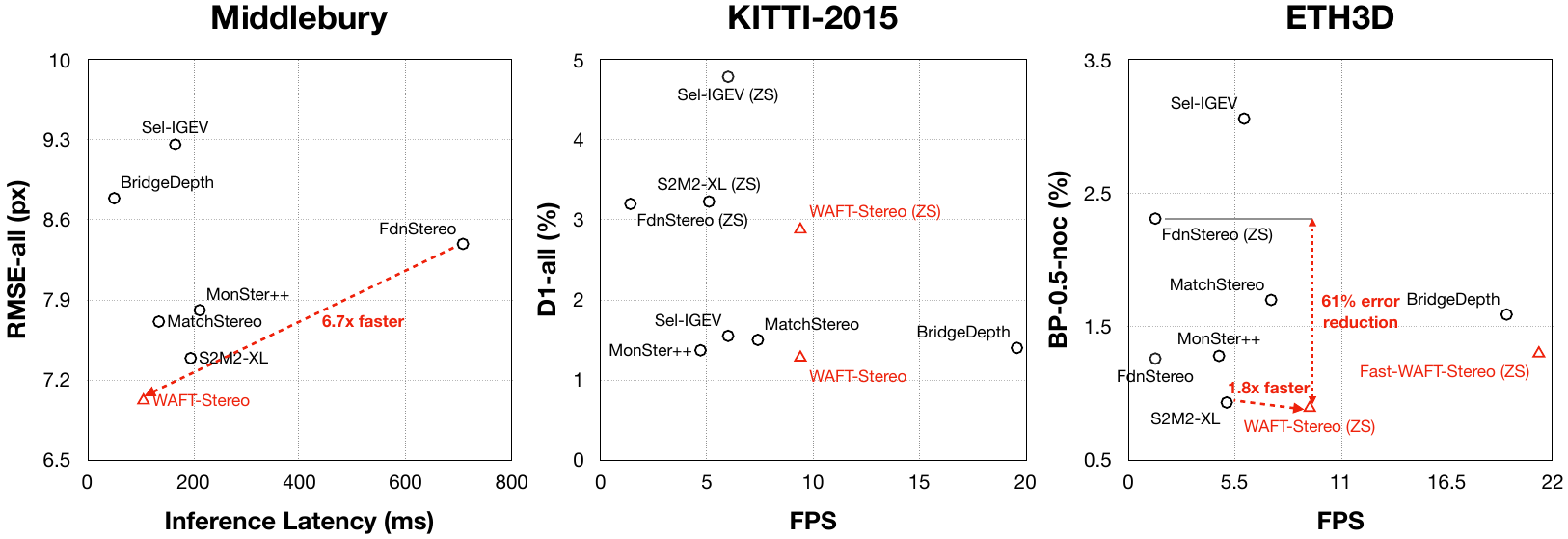}
    \caption{WAFT-Stereo achieves state-of-the-art performance on Middlebury~\cite{middlebury}, KITTI-2015~\cite{menze2015object}, and ETH3D~\cite{eth3d} public benchmarks. Our best-performing model reduces the zero-shot error on ETH3D by at least 61\%, while being $1.8-6.7\times$ faster than leading methods~\cite{wen2025foundationstereo, min2025s2m2}. Our real-time model can process 540p stereo pairs at 21 FPS, while maintaining competitive performance. `ZS' denotes zero-shot submissions.}
    \label{fig:trade-off}
\end{figure}

\myparagraph{\textBF{Deep Stereo Matching}} Modern stereo matching methods are largely driven by deep learning~\cite{zbontar2015computing, mayer2016large, lipson2021raft, igev, xu2024igevpp, selective_stereo, cheng2025monster++, guan2025bridgedepth, guan2024neural, chen2024mocha, jiang2025defom, min2025s2m2, weinzaepfel2023croco, wen2025foundationstereo, gmstereo, guo2024stereo, karaev2023dynamicstereo, min2025depthfocus}. While several methods~\cite{mayer2016large, weinzaepfel2023croco} formulate stereo matching as a direct dense prediction problem, most leading methods~\cite{cheng2025monster++, min2025s2m2, guan2025bridgedepth, wen2025foundationstereo} adopt the RAFT paradigm~\cite{teed2020raft, lipson2021raft}, iteratively refining predictions by regressing disparity updates.

WAFT-Stereo follows the iterative refinement framework. However, unlike many existing iterative methods~\cite{cheng2025monster++, wen2025foundationstereo, lipson2021raft}, WAFT-Stereo performs one-step classification of discretized disparity before regression-based iterative updates. Prior work has used cost volumes to perform classification because cost volume values can naturally represent visual similarities and therefore matching probabilities after softmax normalization~\cite{min2025s2m2, guan2025bridgedepth, zhao2023high, peng2022rethinking, tankovich2021hitnet}. However, WAFT-Stereo demonstrates that the benefit of classification can be realized just by the classification formulation itself, without using cost volumes or cost-volume-specific designs.  

\myparagraph{\textBF{Cost Volume \& Warping}} Cost volumes originated from classical stereo matching methods~\cite{hosni2012fast, scharstein2002taxonomy}, and have become a standard component in modern optical flow and stereo matching methods~\cite{zbontar2015computing, sun2018pwc, wen2025foundationstereo, lipson2021raft, cheng2025monster++, min2025s2m2}. Many recent works~\cite{wen2025foundationstereo, min2025s2m2, guan2025bridgedepth, cheng2025monster++, teed2020raft, lipson2021raft} focus on stronger cost-volume-specific designs for stereo matching.

Warping, in contrast, was primarily studied in the context of optical flow~\cite{brox2004high, memin1998multigrid, black1996robust}. Warping is similar, although not identical, to a special case of partial cost volumes~\cite{sun2018pwc} with look-up radius 1. Despite its simplicity, warping-based designs received much less attention in the past several years, largely due to the strong empirical performance of cost volumes. Recently, WAFT~\cite{wang2025waft} revisited the warping formulation in optical flow. It demonstrated that purely warping-based methods can surpass cost-volume-based methods in both accuracy and efficiency.

WAFT-Stereo builds on the warping framework introduced in WAFT~\cite{wang2025waft}. It adopts a fully warping-based design, removing the reliance on cost-volume-specific network designs used in leading stereo matching methods~\cite{wen2025foundationstereo, min2025s2m2, cheng2025monster++, lipson2021raft, guan2025bridgedepth, jiang2025defom}. This simple warping framework uses standard off-the-shelf architecture components and provides strong performance along with high efficiency.

\begin{figure}[t]
    \centering
    \includegraphics[width=1.0\linewidth]{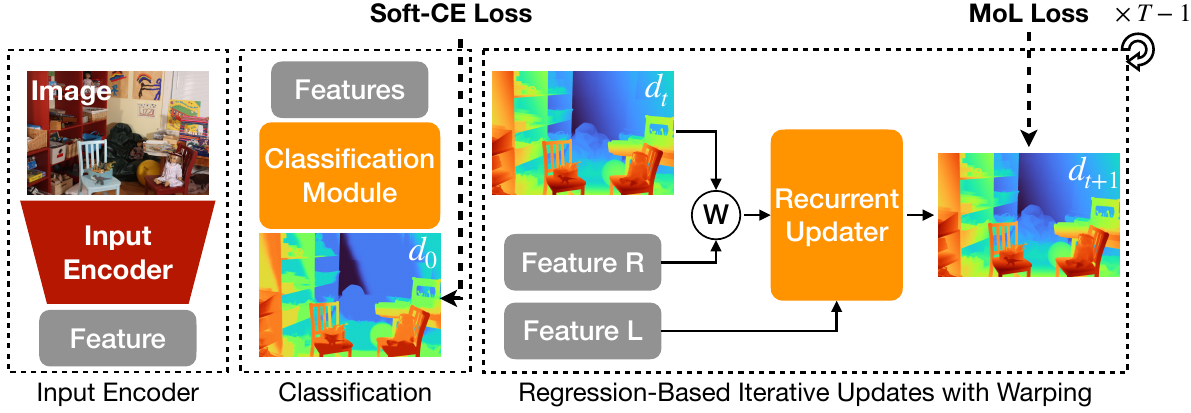}
    \caption{WAFT-Stereo consists of three parts: (1) an input encoder that extracts features from images; (2) a classification step that estimates probabilities over preset disparity bins, supervised by a soft-cross-entropy loss; and (3) a recurrent updater that takes backward-warped right view features as input and regresses disparity updates, supervised by a Mixture-of-Laplace loss~\cite{wang2024sea} for $T-1$ steps.}
    \label{fig:arch}
\end{figure}

\myparagraph{\textBF{Stereo Matching Data}} High-quality training data is critical for modern stereo matching methods. However, annotating real-world stereo pairs with accurate ground-truth disparity is expensive and technically challenging. As a result, the total size of publicly available real-world datasets~\cite{bao2020instereo2k, ramirez2023booster, middlebury, geiger2012we, menze2015object} is three orders of magnitude smaller than synthetic datasets~\cite{sceneflow2016, wen2025foundationstereo, tartanair, raistrick2024infinigen, yang2019drivingstereo, fallingthings, raistrick2023infinite, yan2025makes, crestereo, sintel, cabon2020virtual, Mehl2023_Spring, tosi2021smd, patel2025tartanground}. To mitigate the sim-to-real gap, most leading methods adopt mixed training strategies that combine synthetic and real data.

In contrast, WAFT-Stereo is able to outperform leading approaches on public leaderboards when trained exclusively on synthetic data, highlighting its strong generalization capability.

\section{Method}

In this section, we first review the warping-based iterative refinement in WAFT~\cite{wang2025waft}, explain why it is advantageous compared to the cost-volume counterpart, and describe its implementation in WAFT-Stereo. We then introduce additional technical components, including the proposed classification step.

\subsection{Iterative Refinement with Warping}

\begin{figure}[t]
    \centering
    \includegraphics[width=1.0\linewidth]{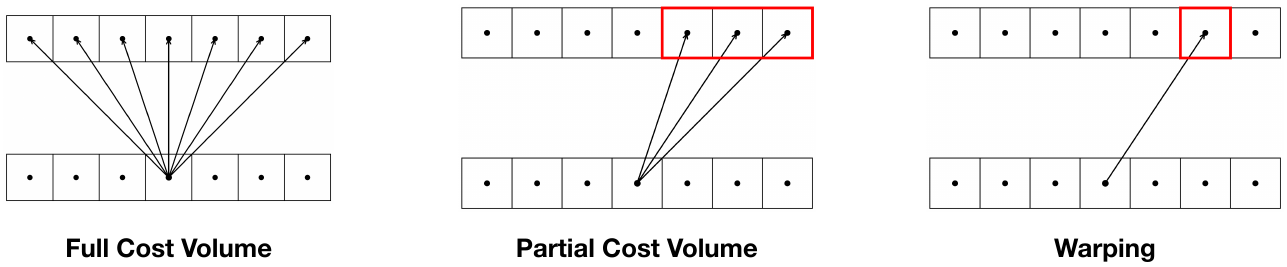}
    \caption{Full cost volumes compute matching costs for all disparity candidates; partial cost volumes compute costs only in a small window around the current disparity estimate; warping aligns the target feature using the current estimate and concatenates aligned and reference features, without computing the matching cost.}
    \label{fig:warping}
\end{figure}

Given a rectified stereo pair $I_L, I_R \in \mathbb{R}^{H\times W\times 3}$, stereo matching aims to estimate a dense disparity field $d \in \mathbb{R}^{H\times W}$ that maps each pixel in $I_L$ to its corresponding location in $I_R$ along the horizontal epipolar line. Modern iterative methods~\cite{wen2025foundationstereo, cheng2025monster++, min2025s2m2} typically extract image features, construct a cost volume at reduced resolution (commonly $1/4$ scale), and recurrently refine the disparity by indexing into the precomputed cost volume.

In optical flow, WAFT~\cite{wang2025waft} replaces cost-volume indexing with feature-space warping: at each iteration, it backward-warps the target-view feature map using the current flow estimate and feeds the aligned features into the recurrent update. This mechanism transfers naturally to stereo matching, since disparity can be viewed as the 1D horizontal component of optical flow.

Let $F(I_L), F(I_R) \in \mathbb{R}^{h\times w\times c}$ denote the extracted feature maps, $d_{\mathrm{cur}} \in \mathbb{R}^{h\times w}$ be the current disparity estimate at the same resolution. The backward warping operator $\texttt{Warp}: \mathbb{R}^{h\times w} \to \mathbb{R}^{h\times w\times c}$ applied to $F(I_R)$ is defined as
\[
\texttt{Warp}(d_{\mathrm{cur}})_p = F(I_R)_{(p_h,\; p_w - (d_{\mathrm{cur}})_p)},
\]
where $p=(p_h,p_w)$ is a pixel in the left view. In practice, $\texttt{Warp}(d_{\mathrm{cur}})$ is computed via bilinear sampling and concatenated with $F(I_L)$ as input to the next refinement iteration.

Warping offers two key advantages over cost volumes. First, its computation and memory scale linearly with the spatial resolution with no dependency on the diparity range, enabling high-resolution indexing~\cite{wang2025waft} that improves accuracy. Second, warping eliminates the need for cost-volume-specific designs, which are typically computationally expensive~\cite{igev, wen2025foundationstereo, cheng2025monster++} in practice. Benefiting from a simpler and more standard network design, WAFT-Stereo can process 540p input at 10 FPS, representing a $6.7\times$ speedup over FoundationStereo~\cite{wen2025foundationstereo} and a $1.8\times$ speedup over S2M2-XL~\cite{min2025s2m2}.

\subsection{Classification before Regression}

\begin{figure}[t]
    \centering
    \includegraphics[width=1.0\linewidth]{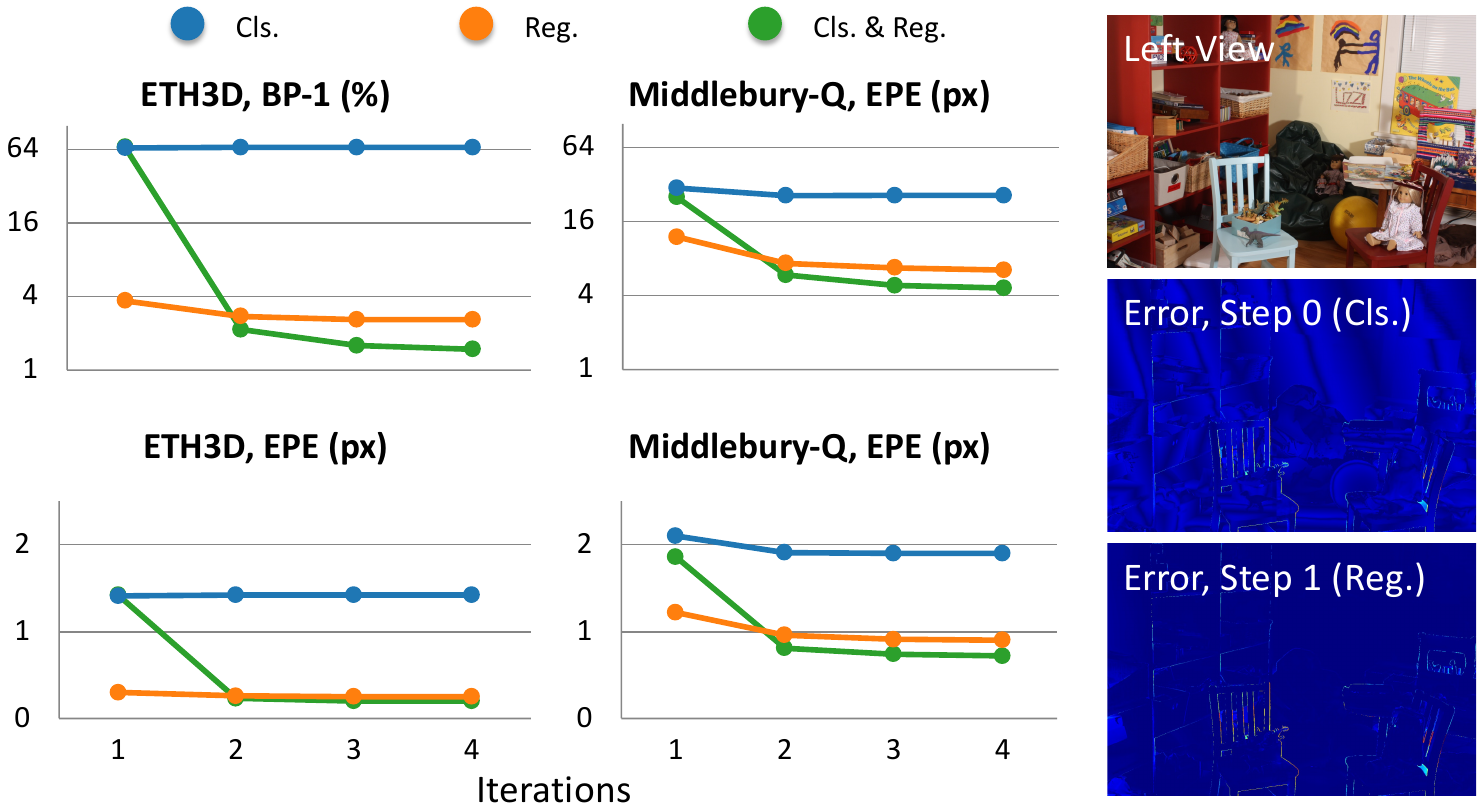}
    \caption{\textBF{Left}: combining classification and regression achieves better performance than using either one alone. \textBF{Right}: the classification step provides a rough estimate, which is later refined by regressions.}
    \label{fig:cls-reg}
\end{figure}

In optical flow, many iterative methods~\cite{wang2025waft, wang2024sea, huang2022flowformer, morimitsu2025dpflow, jahedi2024ccmr} start from an all-zero flow field and recurrently regress residual updates. This strategy works well because typical flow magnitudes are small between consequent frames. In stereo matching, however, large displacements that span hundreds of pixels are common, making purely regression-based iterative updates harder to predict and slower to converge. This is consistent with our observation that regression-only iterative stereo methods~\cite{wen2025foundationstereo, cheng2025monster++, xu2024igevpp} often require a relatively large number of refinement iterations at inference time (e.g., 32), which degrades efficiency.

Classification is known to be more effective than regression in many vision tasks. In stereo matching, cost volumes naturally represent matching probabilities over disparity candidates by computing the feature similarity. Several existing methods~\cite{zhao2023high, guan2025bridgedepth, min2025s2m2} have performed this classification using cost volumes: BridgeDepth~\cite{guan2025bridgedepth} directly supervises the softmax-normalized cost volume with a cross-entropy loss, and derives the initial disparity estimate via soft argmax. S2M2~\cite{min2025s2m2} applies an optimal transport algorithm based on cost volumes to obtain an initial disparity estimate, supervised by the Probabilistic Mode Concentration (PMC) loss. These methods typically require fewer iterations (e.g., 4) in inference than regression-only iterative methods.

WAFT-Stereo shows that the key reason for faster convergence is the classification formulation itself rather than the use of cost volumes or the associated specialized designs. In particular, our classifier can share the same network design as the regression counterpart (a standard vision transformer~\cite{dosovitskiy2020image}), while operating on warped features instead of cost volumes.

In the following, we introduce our implementation of the classification step. Denote the maximum disparity as $D_{max}$, the number of disparity bins as $B$, the ground-truth disparity as $d_{gt}\in \mathbb{R}^{H\times W}$. We define the soft target distribution $P_{gt}$ for each pixel $p$ in $I_L$ as:
\[
(P_{gt})_{p} = \text{Softmax}(\left[-\left|(d_{gt})_p - i \times \frac{D_{max}}{B-1}\right|, \quad i=0,\ldots,B-1\right])
\]
WAFT-Stereo predicts the distribution $P$ in its first step and supervises it with the soft cross-entropy loss (equivalently, KL divergence up to a constant):
\[
(L_{cls})_{p}=-\sum_{i=0}^{B-1}{(P_{gt})_{p, i}\log{P_{p, i}}}
\]

As shown in \Cref{tab:ablation}, in the same 4-iteration setting, replacing the first regression-based iteration with classification substantially improves accuracy with negligible additional cost. \Cref{fig:cls-reg} further illustrates the mechanism: the initial classification provides a coarse but stable estimate that is subsequently refined by the regression-based updates, yielding better final accuracy than the regression-only variant. Although regression can have a better initialization on average, it gets some pixels (those with large displacements) very wrong and will be stuck for them. Classification plus regression solves this issue.

We show additional results on ETH3D (train) to support the claims above. The first step of ``Cls.\& Reg'' achieves 0.89 EPE for pixels with 40px+ disparity and 1.80 EPE for 50px+, better than 1.11 EPE and 2.88 EPE from the first step of ``Reg''.

\subsection{Implementation}

As shown in~\Cref{fig:arch}, the overall architecture of WAFT-Stereo largely follows WAFT~\cite{wang2025waft}, with an additional decoder for the initial classification step. Below we describe each component, highlighting the simplifications and improvements over the original WAFT design.

\myparagraph{\textBF{Input Encoder}} The input encoder $F$ extracts features from the rectified stereo pair $(I_L, I_R)$. Given a pretrained backbone, WAFT-Stereo freezes the backbone weights and fine-tunes the model using low-rank adapters (LoRA)~\cite{hu2022lora}, instead of attaching a side-tunable small U-Net as in WAFT. We use a fully trainable DPT head~\cite{ranftl2021vision} to upsample low-resolution features. We ablate different backbone choices~\cite{yang2024depth, simeoni2025dinov3, wang2025pi} in \Cref{tab:ablation,tab:ablation_data}. Following leading stereo methods~\cite{wen2025foundationstereo, guan2025bridgedepth, jiang2025defom, cheng2025monster++}, we use DepthAnythingV2-L~\cite{yang2024depth} for all benchmark submissions.

\myparagraph{\textBF{Regression-Based Recurrent Updater}} At iteration $t$, the recurrent updater takes the left-view feature $F(I_L)$, the warped right-view feature $\texttt{Warp}(d_t)$, and the hidden state $h_t\in\mathbb{R}^{h\times w\times c}$ as input, and outputs the next hidden state $h_{t+1}$. We adopt the same overall design as WAFT: a ViT-small~\cite{dosovitskiy2020image} followed by a DPT~\cite{ranftl2021vision} feature upsampler. Compared to WAFT, we replace the high-resolution skip connection between hidden states with ResNet blocks, which substantially improves accuracy (see \Cref{tab:ablation}).

\myparagraph{\textBF{Classification Module}} The classification module shares the same architecture as the recurrent updater (ViT-S + DPT). Given the predicted disparity distribution $P$, we compute the initial disparity estimate via soft-argmax and use it to perform the first warping operation in the regression stage:
\[
d_{0}=\sum_{i=0}^{B-1}\left(i\cdot \frac{D_{\max}}{B-1}\right) P_{i}.
\]

\myparagraph{\textBF{Prediction Head \& Loss}} We use the Mixture-of-Laplace (MoL) loss~\cite{wang2024sea} for the regression-based iterative updates. The initial distribution $P$ and the MoL parameters are predicted from the hidden state $h_t$ using an MLP, and are then upsampled to the input resolution via convex upsampling~\cite{teed2020raft}. The total training objective sums the classification loss and the discounted MoL losses across iterations:
\[
L_{\mathrm{total}} = L_{\mathrm{cls}} + \sum_{i=1}^{T-1}\gamma^{T-i} L_{\mathrm{MoL}}^{i},
\]
where $T$ is the total number of iterations (including the initial classification step), $\gamma$ is a decay factor, and $L_{\mathrm{MoL}}^{i}$ is the MoL loss at regression iteration $i$.

\myparagraph{\textBF{Improvements over Existing Methods}} WAFT-Stereo can be viewed as a meta-architecture that simplifies the network design for stereo matching. It combines a lightweight warping operator with standard building blocks (e.g., ViTs and ResNet-style modules), allowing it to leverage well-optimized kernels while avoiding the substantial memory overhead of 3D cost volumes~\cite{wen2025foundationstereo, igev}. Moreover, the accuracy of WAFT-Stereo consistently improves with larger backbones and increased training data, highlighting its strong scalability.

\section{Experiments}

\setlength\tabcolsep{.4em}
\begin{table*}[t]
    \centering
    \resizebox{1.0\linewidth}{!}{
    \begin{tabular}{lccccccc}
    \toprule
     \multirow{2}{*}[\multirowcenter]{Method} & \multirow{2}{*}[\multirowcenter]{Zero-Shot} &\multicolumn{6}{c}{ETH3D (test)} \\
     \cmidrule(l{0.5ex}r{0.5ex}){3-8} 
     & & BP-0.5-noc$\downarrow$ & BP-1-noc$\downarrow$ & BP-2-noc$\downarrow$ & BP-0.5-all$\downarrow$ & BP-1-all$\downarrow$ & BP-2-all$\downarrow$ \\
     \midrule
     RAFT-Stereo~\cite{teed2020raft} & $\text{\ding{55}}$ & 7.04 & 2.44 & 0.44 & 7.33 & 2.60 & 0.56 \\
     CREStereo~\cite{crestereo} & $\text{\ding{55}}$ & 3.58 & 0.98 & 0.22 & 3.75 & 1.09 & 0.29 \\
     CroCo-Stereo~\cite{weinzaepfel2023croco} & $\text{\ding{55}}$ &  3.27 & 0.99 & 0.39 & 3.51 & 1.14 & 0.50 \\
     Selective-IGEV~\cite{selective_stereo} & $\text{\ding{55}}$ & 3.06 & 1.23 & 0.22 &  3.46 & 1.56 & 0.51 \\
     BridgeDepth~\cite{guan2025bridgedepth} & $\text{\ding{55}}$ & 1.59 & 0.38 & 0.11 & 1.76 & 0.50 & 0.16\\
     MonSter++~\cite{cheng2025monster++} & $\text{\ding{55}}$ & 1.28 & 0.25 & 0.09 & 1.59 & 0.45 & 0.23\\
     MatchStereo~\cite{yan2025matchattention} & $\text{\ding{55}}$ & 1.70 & 0.70 & 0.17 & 1.90 & 0.79 & 0.20\\
     DEFOM-Stereo~\cite{jiang2025defom} & $\text{\ding{55}}$ & 2.21 & 0.70 & 0.14 & 2.43 & 0.78 & 0.19\\
     FoundationStereo~\cite{wen2025foundationstereo} & $\text{\ding{55}}$ & 1.26 & 0.26 & 0.08 & 1.56 & 0.48 & 0.26 \\
     S2M2-XL~\cite{min2025s2m2} & $\text{\ding{55}}$ & 0.93 & \textBF{0.22} & \textBF{0.06} & 1.03 & \textBF{0.26} & 0.08 \\
     \midrule
     FoundationStereo~\cite{wen2025foundationstereo} & $\text{\ding{51}}$ & 2.31 & 1.52 & - & - & - & - \\
     Ours (DAv2S, 4) & $\text{\ding{51}}$ & 1.30 & 0.31 & 0.11 & 1.42 & 0.36 & 0.13\\
     Ours (DAv2L, 5) & $\text{\ding{51}}$ & \textBF{0.89} & \textBF{0.28} & 0.07 & \textBF{0.97} & 0.31 & \textBF{0.07}\\
    \bottomrule
    \end{tabular}
    }
    \caption{Trained on synthetic data, WAFT-Stereo achieves the best BP-0.5 and BP-2-all on ETH3D benchmark, demonstrating its strong sim-to-real generalization. This also represents a 61\%--81\% error reduction on the best established zero-shot results.}
    \label{tab:eth3d}
\end{table*}

In this section, we describe the training pipeline of WAFT-Stereo, analyze the results of benchmark submissions, and ablate our design choices, including the training data we used.

\myparagraph{\textBF{Training Pipeline \& Datasets}} WAFT-Stereo is first trained on synthetic data for zero-shot evaluation and submissions, then fine-tuned on real-world data for final submissions. The synthetic datasets we used are SceneFlow~\cite{sceneflow2016}, FallingThings~\cite{fallingthings}, FSD~\cite{wen2025foundationstereo}, TartanAir~\cite{tartanair}, TartanGround~\cite{patel2025tartanground}, Spring~\cite{Mehl2023_Spring}, CREStereo~\cite{crestereo}, Sintel~\cite{sintel}, Virtual KITTI 2~\cite{cabon2020virtual}, UnrealStereo4K~\cite{tosi2021smd}, WMGStereo~\cite{yan2025makes}, and HR-VS~\cite{yang2019hierarchical}. The real-world datasets we used for fine-tuning are Booster~\cite{ramirez2023booster}, InStereo2K~\cite{bao2020instereo2k}, KITTI~\cite{menze2015object, geiger2012we}, and Middlebury~\cite{middlebury}.

We report results on ETH3D~\cite{eth3d}, KITTI-2012/2015~\cite{geiger2012we, menze2015object}, and Middlebury~\cite{middlebury} public benchmarks, and conduct ablations on the training splits of ETH3D~\cite{eth3d}, KITTI~\cite{geiger2012we, menze2015object}, and Middlebury~\cite{middlebury}.

\myparagraph{\textBF{Metrics}} We report commonly used evaluation metrics, including (1) \textBF{BP-X}: the percentage of pixels whose error exceeds X pixels, (2) \textBF{D1}: the percentage of pixels whose error exceeds 3 pixels and 5\% of the ground-truth disparity, and (3) \textBF{RMSE}: the root mean square error.

\myparagraph{\textBF{Implementation Details}} Following WAFT~\cite{wang2025waft}, we perform warping at half resolution, i.e., $(h,w)=(\tfrac{H}{2},\tfrac{W}{2})$, and apply an $8\times 8$ patchifier before the ViT blocks in the recurrent updater. We use 4 ResNet blocks for high-resolution processing within the updater. For the input encoder, we set the LoRA~\cite{hu2022lora} rank to 8. We use $B=40$ preset disparity bins for classification with a maximum disparity $D_{max}=800$.

For brevity, we denote a WAFT-Stereo configuration by its pretrained backbone and the number of iterations, e.g., (DAv2-L, 5).

\subsection{Training on Synthetic Data}
\label{sec:syn}

In the first stage, we train WAFT-Stereo exclusively on a mixture of synthetic stereo datasets~\cite{patel2025tartanground, tartanair, sceneflow2016, yan2025makes, cabon2020virtual, crestereo, wen2025foundationstereo, sintel, fallingthings, Mehl2023_Spring, tosi2021smd, yang2019hierarchical}. This training mixture (denoted as \texttt{SynLarge}) contains approximately 7.7 million stereo pairs in total. We train on 480p random crops with batch size 32 and learning rate $5\times 10^{-4}$ for 400k steps, using AdamW optimizer~\cite{loshchilov2017decoupled} with OneCycle scheduler~\cite{smith2019super}.

Unless otherwise specified, we use this synthetic-only checkpoint for all zero-shot evaluations and benchmark submissions. As we will show later, synthetic-only training already yields strong performance and strong sim-to-real generalization for WAFT-Stereo.

\subsection{ETH3D}

We report zero-shot results on ETH3D public benchmark using the model trained exclusively on synthetic data (see \Cref{sec:syn}). We follow the benchmark protocols and report BP-0.5, BP-1, and BP-2 on both all pixels and non-occluded pixels (denoted as `all' and `noc' suffixes).

As shown in~\Cref{tab:eth3d}, WAFT-Stereo ranks first on BP-0.5 and BP-2-all among all public submissions, demonstrating its high accuracy. WAFT-Stereo is also more efficient than leading methods~\cite{wen2025foundationstereo, min2025s2m2, cheng2025monster++}. Compared to FoundationStereo~\cite{wen2025foundationstereo}, WAFT-Stereo achieves a $6.7\times$ lower latency and uses $5.2\times$ fewer MACs. Compared to S2M2-XL~\cite{min2025s2m2}, WAFT-Stereo achieves $1.8\times$ lower latency and uses $2.7\times$ fewer MACs. Compared to MonSter++~\cite{cheng2025monster++}, WAFT-Stereo achieves $2.0\times$ lower latency and uses $2.1\times$ fewer MACs.

WAFT-Stereo also demonstrates strong sim-to-real generalization. Compared to the strongest established zero-shot baseline, FoundationStereo~\cite{wen2025foundationstereo}, our best-performing model reduces BP-0.5 by 61\% and BP-1 by 81\% on non-occluded pixels. Moreover, our fastest variant, built on the DepthAnythingV2-S backbone~\cite{yang2024depth}, achieves 44\% lower BP-0.5 and 80\% lower BP-1 on non-occluded pixels, while maintaining real-time speed, processing 540p stereo pairs at more than 21 FPS.
\setlength\tabcolsep{.4em}
\begin{table*}[t]
    \centering
    \resizebox{1.0\linewidth}{!}{
    \begin{tabular}{lccccccc}
    \toprule
     \multirow{2}{*}[\multirowcenter]{Method} & \multirow{2}{*}[\multirowcenter]{Zero-Shot} & \multicolumn{2}{c}{KITTI-12 (test)} & \multicolumn{4}{c}{KITTI-15 (test)} \\
     \cmidrule(l{0.5ex}r{0.5ex}){3-4} \cmidrule(l{0.5ex}r{0.5ex}){5-8}
     & & BP-2-noc$\downarrow$ & BP-2-all$\downarrow$ & D1-bg-noc $\downarrow$ & D1-noc $\downarrow$ & D1-bg-all $\downarrow$ & D1-all$\downarrow$ \\
     \midrule
     GWCNet~\cite{guo2019group} & $\text{\ding{55}}$ & 2.16 & 2.71 & 1.61 & 1.92 & 1.74 & 2.11 \\
     RAFT-Stereo~\cite{lipson2021raft} & $\text{\ding{55}}$ & 1.44 & 1.69 & 1.58 &  1.82 & 1.92 & 2.42 \\
     CREStereo~\cite{crestereo} & $\text{\ding{55}}$ & 1.72 & 2.18 & 1.33 & 1.54 & 1.45 & 1.69\\
     DLNR~\cite{zhao2023high} & $\text{\ding{55}}$ & - & - & 1.42 & 1.61 & 1.60 & 1.76 \\
     MoCha-Stereo~\cite{chen2024mocha} & $\text{\ding{55}}$ & 1.51 & 1.91 & 1.17 & 1.40 & 1.27 & 1.49 \\
     Selective-IGEV~\cite{selective_stereo} & $\text{\ding{55}}$ & 1.59 & 2.05 & 1.22 & 1.44 & 1.33 & 1.55 \\
     Croco-Stereo~\cite{weinzaepfel2023croco} & $\text{\ding{55}}$ & - & - &  1.30 & 1.51 & 1.38 & 1.59 \\
     NMRF~\cite{zhao2023high} & $\text{\ding{55}}$ & 1.59 & 2.07 & 1.18 & 1.46 & 1.28 & 1.57 \\
     BridgeDepth~\cite{guan2025bridgedepth} & $\text{\ding{55}}$ & 1.32 & 1.65 & 1.05 & 1.31 & 1.13 & 1.40 \\
     DEFOM-Stereo~\cite{jiang2025defom} & $\text{\ding{55}}$ & 1.43 & 1.79 & 1.15 & 1.33 & 1.25 & 1.41 \\
     MatchStereo~\cite{yan2025matchattention} & $\text{\ding{55}}$ & 1.63 & 2.08 & 1.23 & 1.40 & 1.34 & 1.50 \\
     MonSter++~\cite{cheng2025monster++} & $\text{\ding{55}}$ & 1.30 & 1.70 & 1.02 & 1.29 & 1.12 & 1.37\\
     Ours (DAv2L, 5) & $\text{\ding{55}}$ & \textBF{1.18} & \textBF{1.47} & \textBF{0.98} & \textBF{1.21} & \textBF{1.05} & \textBF{1.28} \\
     \midrule 
     Selective-IGEV~\cite{selective_stereo} & $\text{\ding{51}}$ & - & - & 2.91 & 4.57 & 3.06 & 4.79 \\
     FoundationStereo~\cite{wen2025foundationstereo} & $\text{\ding{51}}$ & - & - & 2.48 & 3.08 & 2.59 & 3.20 \\
     S2M2-XL~\cite{min2025s2m2} & $\text{\ding{51}}$ & - & - & 2.51 & 3.12 & 2.61 & 3.23 \\ 
     Ours (DAv2L, 5) & $\text{\ding{51}}$ & - & - & \textBF{2.39} & \textBF{2.79} & \textBF{2.47} & \textBF{2.88} \\
    \bottomrule
    \end{tabular}
    }
    \caption{WAFT-Stereo ranks first on KITTI public benchmarks, reducing BP-2 by 13\% on KITTI-2012 and D1 by 6\% on KITTI-2015. It also outperforms leading methods when not fine-tuned on the training split, reducing D1 by 9\%.}
    \label{tab:kitti}
\end{table*}

\subsection{KITTI-2012/2015}

We report both fine-tuned and zero-shot results on KITTI. For fine-tuned submissions, we further train the synthetic-pretrained model on the KITTI training splits using $372\times1240$ crops for 3k steps, with batch size 16 and learning rate $10^{-4}$. We follow the standard evaluation protocols, report D1 for KITTI-2015 and BP-2 for KITTI-2012, each on both all pixels and non-occluded pixels.

As shown in \Cref{tab:kitti}, WAFT-Stereo achieves the best BP-2 on KITTI-2012 and the best D1 on KITTI-2015, representing a 13\% and a 6\% error reduction on previous state-of-the-art method, MonSter++~\cite{cheng2025monster++}, respectively. We achieve this with a $2.0\times$ speed-up and the same DepthAnythingV2-L~\cite{yang2024depth} backbone.

Since prior work has observed strong dataset bias on KITTI~\cite{min2025s2m2}, we additionally report zero-shot performance on KITTI-2015. Trained exclusively on synthetic data, WAFT-Stereo achieves the best zero-shot result, reducing D1 by at least 10\% compared to previous leading methods~\cite{selective_stereo, wen2025foundationstereo, min2025s2m2}.

\subsection{Middlebury}

\setlength\tabcolsep{.4em}
\begin{table*}[t]
    \centering
    \resizebox{1.0\linewidth}{!}{
    \begin{tabular}{lcccccc}
    \toprule
     \multirow{2}{*}[\multirowcenter]{Method} &\multicolumn{6}{c}{Middlebury (test)} \\
     \cmidrule(l{0.5ex}r{0.5ex}){2-7} 
     & BP-2-noc$\downarrow$ & BP-2-all$\downarrow$ & BP-4-noc$\downarrow$ & BP-4-all$\downarrow$ & RMSE-noc$\downarrow$ & RMSE-all$\downarrow$ \\
     \midrule
     RAFT-Stereo~\cite{lipson2021raft} & 4.74 & 9.37 & 2.75 & 6.42 & 8.41 & 12.6 \\
     CREStereo~\cite{crestereo} & 3.71 & 8.13 & 2.04 & 5.05 & 7.70 & 10.5 \\
     CroCo-Stereo~\cite{weinzaepfel2023croco} & 7.29 & 11.1 & 4.18 & 6.75 & 8.91 & 10.6 \\
     GMStereo~\cite{gmstereo} & 7.14 & 11.7 & 2.96 & 6.07 & 6.45 & 8.03 \\
     MoCha-Stereo~\cite{chen2024mocha} & 3.51 & 6.74 & 2.34 & 4.54 & 7.69 & 9.91 \\
     Selective-IGEV~\cite{selective_stereo} & 2.51 & 6.04 & 1.36 & 3.74 & 7.26 & 9.26 \\
     MonSter++~\cite{cheng2025monster++} & 2.60 & 6.40 & 1.18 & 3.52 & 6.03 & 7.81 \\
     DEFOM-Stereo~\cite{jiang2025defom} & 2.39 & 5.02 & 1.22 & 3.02 & 5.81 & 7.73 \\
     BridgeDepth~\cite{guan2025bridgedepth} & 3.78 & 6.92 & 1.61 & 3.48 & 7.16 & 8.79 \\
     MatchStereo~\cite{yan2025matchattention} & 1.85 & 4.90 & 0.91 & 3.02 & 5.94 & 7.71 \\
     FoundationStereo~\cite{wen2025foundationstereo} & 1.84 & 4.26 & 1.04 & 2.72 & 6.48 & 8.39\\
     S2M2-XL~\cite{min2025s2m2} & \textBF{1.15} & \textBF{2.94} & \textBF{0.54} & \textBF{1.63} & 6.01 & 7.39 \\
     Ours (DAv2L, 5)  & 2.53 & 4.39 & 1.05 & 2.15 & \textBF{5.61} & \textBF{7.02} \\
    \bottomrule
    \end{tabular}
    }
    \caption{WAFT-Stereo achieves the best RMSE on Middlebury public benchmark, reducing the error by 6\% on non-occluded pixels and 5\% on all pixels.}
    \label{tab:middlebury}
\end{table*}

We report fine-tuned results on this benchmark. Starting from the synthetic-pretrained checkpoint, we fine-tune WAFT-Stereo on a mixture of real-world and synthetic data~\cite{wen2025foundationstereo, fallingthings, crestereo, Mehl2023_Spring, tosi2021smd, yang2019hierarchical, bao2020instereo2k, ramirez2023booster, middlebury} using $540\times960$ crops for 50k steps, with batch size 16 and learning rate $2\times10^{-4}$. We follow the standard protocol and report BP-2, BP-4, and RMSE on both all pixels and non-occluded pixels.

As shown in~\Cref{tab:middlebury}, WAFT-Stereo achieves the best RMSE, improving over previous methods by 5\%. With the same DepthAnythingV2-L~\cite{yang2024depth} backbone, it matches or surpasses FoundationStereo~\cite{wen2025foundationstereo}, MonSter++~\cite{cheng2025monster++}, and DEFOM-Stereo~\cite{jiang2025defom} on BP-4 while being $6.7\times$, $2.0\times$, and $1.5\times$ faster, respectively.

However, WAFT-Stereo falls behind the leading methods~\cite{min2025s2m2, wen2025foundationstereo, yan2025matchattention} in terms of the averaged BP-2 metric across all 15 scenes. We observe that this performance gap is largely dominated by a single particularly challenging scene, \texttt{Classroom2E}, where the stereo pair exhibits substantial illumination differences. Excluding this scene significantly narrows the gap.

Despite this, WAFT-Stereo remains on the Pareto frontier when considering accuracy–efficiency trade-offs. Moreover, when performing scene-wise analysis, WAFT-Stereo achieves the best BP-2-all on four, the best BP-2-noc on three, the best BP-4-all on six, and the best BP-4-noc on three out of the total 15 scenes, demonstrating its strong competitiveness.

\subsection{Ablations on Training Data}

\setlength\tabcolsep{.2em}
\begin{table*}[t]
    \centering
    \resizebox{1.0\linewidth}{!}{
    \begin{tabular}{llccccccccc}
    \toprule
     \multirow{2}{*}[\multirowcenter]{Data} & \multirow{2}{*}[\multirowcenter]{Method} & \multicolumn{2}{c}{KITTI-12 (train)} & \multicolumn{2}{c}{KITTI-15 (train)} & \multicolumn{2}{c}{ETH3D (train)} & \multicolumn{2}{c}{Middlebury-Q (train)} & \multirow{2}{*}[\multirowcenter]{Latency}\\
     \cmidrule(l{0.5ex}r{0.5ex}){3-4} \cmidrule(l{0.5ex}r{0.5ex}){5-6} \cmidrule(l{0.5ex}r{0.5ex}){7-8} \cmidrule(l{0.5ex}r{0.5ex}){9-10}
     & & D1-all$\downarrow$ & D1-noc$\downarrow$ & D1-all$\downarrow$ & D1-noc$\downarrow$ & BP-1-all$\downarrow$ & BP-1-noc$\downarrow$ & BP-2-all$\downarrow$ & BP-2-noc$\downarrow$\\
    \midrule
         \multirow{2}{*}{FSD-ZS}& FoundationStereo~\cite{wen2025foundationstereo} & - & 2.30 & - & 2.80 & - & 0.50 & - & 1.30 & 708ms\\
         & Ours (DAv2L, 5) & 2.54 & 2.36 & 2.93 & 2.85 & 0.60 & 0.52 & 2.29 & 1.38 & 106ms\\
        \midrule
       FSD-ZS+ & MonSter++~\cite{cheng2025monster++} & 2.95 & 2.69 & 3.25 & 3.16 & 1.08 & 0.89 & 4.19 & 2.93 & 212ms\\
    \midrule
        \multirow{4}{*}{SynLarge} & BridgeDepth~\cite{guan2025bridgedepth}$^{\dagger}$ & 2.81 & 2.57 & 3.15 & 3.06 & 0.79 & 0.71 & 3.29 & 1.90 & 51ms\\
        & Ours (DAv2S, 4) & 2.80 & 2.58 & 3.20 & 3.08 & 0.86 & 0.76 & 4.14 & 2.66 & 47ms\\
        & Ours (DAv2B, 4) & 2.58 & 2.38 & 2.88 & 2.80 & 0.71 & 0.63 & 2.76 & 1.67 & 66ms\\
        & Ours (DAv2L, 5) & 2.67 & 2.46 & 2.89 & 2.80 & 0.53 & 0.47 & 2.19 & 1.27 & 106ms\\
    \midrule
         \multirow{2}{*}{SceneFlow}& BridgeDepth~\cite{guan2025bridgedepth}$^{\dagger}$ & 3.11 & 2.80 & 4.44 & 4.27 & 1.21 & 1.08 & 4.52 & 3.04 & 51ms\\
         & Ours (DAv2S, 4) & 10.1 & 9.78 & 16.8 & 16.6 & 71.0 & 71.1 & 8.86 & 6.61 & 47ms\\
    \bottomrule
    \end{tabular}
    }
    \caption{We report the zero-shot ablation results on the training splits of KITTI~\cite{menze2015object}, ETH3D~\cite{eth3d} and Middlebury-Q~\cite{middlebury}. The training data is identified in the first column. WAFT-Stereo architecture achieves comparable or better performance than leading methods with the same or fewer training data, while being $2.0-6.7\times$ faster. $\dagger$ denotes our re-implementation of BridgeDepth~\cite{guan2025bridgedepth}, which provides better accuracy than reported in the original paper.}
    \label{tab:ablation_data}
\end{table*}

The strong performance of WAFT-Stereo does not stem merely from larger-scale training data. In \Cref{tab:ablation_data}, we ablate the training data and demonstrate that the WAFT-Stereo architecture achieves comparable or superior accuracy to leading methods~\cite{wen2025foundationstereo, guan2025bridgedepth, cheng2025monster++} under the same or even fewer training data. We report standard metrics on the training splits of KITTI~\cite{geiger2012we, menze2015object}, ETH3D~\cite{eth3d}, and Middlebury~\cite{middlebury}, following existing work~\cite{igev, wen2025foundationstereo, cheng2025monster++, guan2025bridgedepth}.

In the first block, we follow the zero-shot setting of FoundationStereo~\cite{wen2025foundationstereo} (denoted as “FSD-ZS”), restricting the training data to a mixture of synthetic datasets~\cite{sceneflow2016, sintel, crestereo, fallingthings, cabon2020virtual} and a single real-world dataset~\cite{bao2020instereo2k}. Under this setting, WAFT-Stereo achieves performance comparable to FoundationStereo while delivering a $6.7\times$ speed-up and $5.3\times$ fewer MACs.

The second block corresponds to the standard zero-shot setting of MonSter++~\cite{cheng2025monster++}, which uses a larger collection of training datasets~\cite{sceneflow2016, wen2025foundationstereo, tartanair, crestereo, sintel, bao2020instereo2k, karaev2023dynamicstereo, fallingthings, ramirez2023booster, cabon2020virtual, Mehl2023_Spring, wu2022toward, tosi2021smd, yang2019hierarchical, wang2021irs, zhou2019davanet, niklaus20193d} compared to the FSD-ZS setting. Even with fewer data, WAFT-Stereo achieves significantly better performance across all metrics, together with a $2.0\times$ speed-up and $2.1\times$ fewer MACs.

The third block compares WAFT-Stereo with BridgeDepth~\cite{guan2025bridgedepth}, the best zero-shot method under the commonly reported SceneFlow-only training setting~\cite{sceneflow2016}. Using the same DepthAnythingV2-L backbone~\cite{yang2024depth}, WAFT-Stereo significantly outperforms BridgeDepth when trained on 7.7 million diverse synthetic pairs (see \Cref{sec:syn}). Consistently, WAFT-Stereo surpasses BridgeDepth across all public benchmarks~\cite{eth3d, middlebury, menze2015object, geiger2012we} (see \Cref{tab:kitti,tab:middlebury,tab:eth3d}) 

However, as shown in the last block, training WAFT-Stereo solely on SceneFlow (35k pairs) leads to severe overfitting to the dataset bias and consequently degrades generalization. These results suggest that performance under the widely adopted SceneFlow-only zero-shot setting may not faithfully reflect the actual capability of a model.

\subsection{Ablations and Analysis on Architecture Design}
\label{sec:ablation_arch}

We conduct extensive ablations on our architectural design choices in \Cref{tab:ablation}. Unless otherwise specified, all models are trained with 480p image crops, batch size 32, and learning rate $5\times10^{-4}$ for 100k iterations.

We use a smaller synthetic subset~\cite{wen2025foundationstereo, tartanair, crestereo, sceneflow2016} for training in ablation experiments, containing approximately 1.6 million stereo pairs. We report zero-shot performance on the training splits of ETH3D and Middlebury. The maximum disparity is set to $D_{\max}=320$.

\setlength\tabcolsep{.4em}
\begin{table*}[t]
    \centering
    \resizebox{1.0\linewidth}{!}{
    \begin{tabular}{lccccccccc}
    \toprule
     \multirow{2}{*}[\multirowcenter]{Experiment} & \multirow{2}{*}[\multirowcenter]{Feat.} &\multicolumn{2}{c}{\#Iters} & \multirow{2}{*}[\multirowcenter]{\#HR blks} & \multirow{2}{*}[\multirowcenter]{\#Bins} & ETH3D & Middlebury-Q & \multirow{2}{*}[\multirowcenter]{Reg. Loss} & \multirow{2}{*}[\multirowcenter]{\#MACs}\\
     \cmidrule(l{0.5ex}r{0.5ex}){3-4} \cmidrule(l{0.5ex}r{0.5ex}){7-7} \cmidrule(l{0.5ex}r{0.5ex}){8-8}
     & & Cls. & Reg. &  & & BP-1-all$\downarrow$ & BP-2-all$\downarrow$ & \\
     \midrule
     Baseline & DAv2S & 1 & 3 & 4 & 40 & 1.48 & 4.62 & MoL& 560G \\
     \midrule
     \multirow{3}{*}{\textcolor{CadetBlue}{Disp. bins}} & \multirow{3}{*}{DAv2S} & \multirow{3}{*}{1} & \multirow{3}{*}{3} & \multirow{3}{*}{4} & 5 & 2.46 & 6.45 & \multirow{3}{*}{MoL} & \multirow{3}{*}{560G} \\
     & & & & & 20 & 1.70 & 5.04 &\\ 
     & & & & & 80 & 1.27 & 5.39 &\\
     \midrule
     \multirow{3}{*}{\textcolor{orange}{Cls. \& Reg.}} & \multirow{3}{*}{DAv2S} & 0 & 4 & \multirow{3}{*}{4} & \multirow{3}{*}{40} & 2.59 & 6.45 & MoL & 561G \\
     & & 4 & 0 & & & 66.8 & 26.1 & MoL & 560G \\
     & & C.V. & 3 & & & 1.53 & 4.74 & MoL & 561G \\ 
     \midrule
     \textcolor{Mahogany}{HR blks} & DAv2S & 1 & 3 & 0 & 40 & 1.99 & 6.20 & MoL & 494G \\
     \midrule
     \multirow{2}{*}{\textcolor{Plum}{Diff. Feat.}} & DAv2L & 1 & 3 & \multirow{2}{*}{4} & \multirow{2}{*}{40} & 0.80 & 2.63 & MoL & 2172G\\
     & DINOv3L & 1 & 3 & & & 0.87 & 2.80 & MoL & 2175G \\
     \midrule
     \textcolor{teal}{Reg. Loss} & DAv2S & 1 & 3 & 4 & 40 & 2.44 & 6.39 & $L_1$ & 560G \\
    \bottomrule
    \end{tabular}
    }
    \caption{We report the zero-shot ablation results on the training splits of ETH3D and Middlebury-Q.  The effect of changes can be identified through comparisons to the first row. See~\Cref{sec:ablation_arch} for more details.}
    \label{tab:ablation}
\end{table*}

\myparagraph{\textBF{\textcolor{CadetBlue}{Number of Disparity Bins}}}
In the initial classification stage, the number of predefined disparity bins should not be too small. We observe a significant performance drop when only 5 bins are used. Considering the trade-off between accuracy and memory efficiency, we choose $B=40$ as the default setting.

\myparagraph{\textBF{\textcolor{orange}{Classification \& Regression}}}
Combining classification and regression yields the best performance in practice, as shown in~\Cref{fig:cls-reg} and~\Cref{tab:ablation}. Compared to the regression-only variant, the hybrid design consistently achieves better results on both Middlebury and ETH3D.

In contrast, classification primarily provides coarse disparity estimates. The classification-only variant benefits little from iterative refinement and significantly underperforms the other two variants on BP-X metrics.

We also implement a classification variant based on cost volumes, following prior work~\cite{guan2025bridgedepth, zhao2023high}. However, introducing cost volumes does not provide noticeable performance improvements. This observation suggests that the gains mainly stem from the classification formulation itself, rather than from cost-volume-specific design choices.

\myparagraph{\textBF{\textcolor{Mahogany}{High-Resolution Blocks}}} The high-resolution ResNet-blocks~\cite{he2016deep} remarkably improve accuracy. This is reasonable since the low-resolution processing after patchification may lose information about high-resolution details.

\myparagraph{\textBF{\textcolor{Plum}{Different Backbones}}}
WAFT-Stereo can be easily integrated with various pre-trained backbones~\cite{simeoni2025dinov3, yang2024depth}. For fair comparison with prior work, we adopt the DAv2 series~\cite{yang2024depth}, following existing methods~\cite{cheng2025monster++, jiang2025defom, guan2025bridgedepth, wen2025foundationstereo}. We leave more explorations on backbone choices for future work.

\setlength\tabcolsep{.2em}
\begin{table}[t]
    \begin{minipage}[b]{.42\linewidth}
        \centering
        \includegraphics[width=1.0\linewidth]{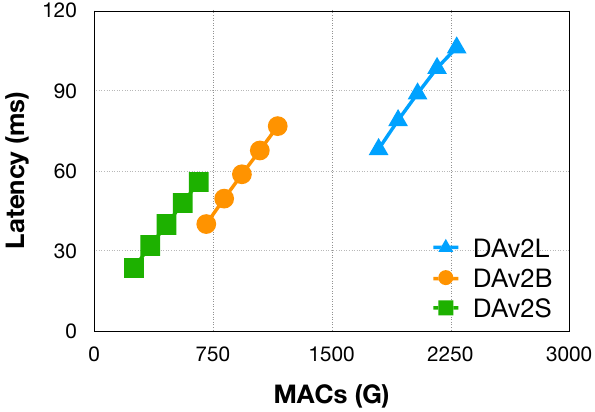}
        \captionof{figure}{The latency grows almost linearly with the number of iterative updates, reflecting reduced parallelism as the iteration count increases.}
        \label{fig:decomp}
    \end{minipage}\hfill
    {\small
    \begin{minipage}[b]{.53\linewidth}
        \centering
        \begin{tabular}{lccc}
        \toprule
        Method & Latency & \#MACs & \#Params \\
        \midrule
            MonSter++~\cite{cheng2025monster++} & 212ms & 4.92T & 0.39B \\
            MatchStereo~\cite{yan2025matchattention} & 135ms & 0.39T & 0.08B \\
            BridgeDepth~\cite{guan2025bridgedepth} & 51ms & 1.34T & 0.35B \\
            DEFOM~\cite{jiang2025defom} & 163ms & 4.91T & 0.38B \\
            S2M2-XL~\cite{min2025s2m2} & 195ms & 6.26T & 0.41B \\
            FdnStereo~\cite{wen2025foundationstereo} & 708ms & 12.1T & 0.38B \\
        \midrule
            Ours (DAv2S, 4) & 47ms & 0.56T & 0.08B \\
            Ours (DAv2B, 4) & 66ms & 1.05T & 0.15B \\
            Ours (DAv2L, 5) & 106ms & 2.29T & 0.38B \\
        \bottomrule
        \end{tabular}
        \captionof{table}{More details on the inference cost of WAFT-Stereo and leading methods, profiled with 540p input, BF16 precision on an L40.}
        \label{tab:cost}
    \end{minipage}
    }
\end{table}

\myparagraph{\textBF{\textcolor{teal}{Regression Loss}}} 
As in optical flow, the Mixture-of-Laplace (MoL) loss~\cite{wang2024sea} yields significantly better performance for iterative refinement than the $L_1$ loss widely adopted in prior work~\cite{lipson2021raft, wen2025foundationstereo, cheng2025monster++, guan2025bridgedepth, min2025s2m2}.

\myparagraph{\textBF{Inference Cost}} 
We provide a detailed cost breakdown in~\Cref{fig:decomp} and~\Cref{tab:cost}. All profiling results use 540p inputs, BF16 precision, and an NVIDIA L40 GPU.

WAFT-Stereo is faster than most leading methods, with BridgeDepth~\cite{guan2025bridgedepth} as the only faster baseline in our comparison. However, WAFT-Stereo remains on the Pareto frontier of the speed--accuracy trade-off (see \Cref{tab:eth3d,tab:kitti,tab:middlebury,tab:ablation_data}).

We further analyze how inference latency grows with an increasing number of iterations.~\Cref{fig:decomp} shows that more iterations reduces parallelism, though WAFT-Stereo already requires substantially fewer iterations than prior iterative methods~\cite{wen2025foundationstereo, igev, lipson2021raft}. We leave better designs for future work.

\section*{Acknowledgements}
This work was partially supported by the National Science Foundation.

\bibliographystyle{splncs04}
\bibliography{main}
\end{document}

%% file: macros.tex
\newcommand{\xpar}[1]{\noindent\textbf{#1}\ \ }
\newcommand{\vpar}[1]{\vspace{3mm}\noindent\textbf{#1}\ \ }

\newcommand{\sect}[1]{Section~\ref{#1}}
\newcommand{\sects}[1]{Sections~\ref{#1}}
\newcommand{\eqn}[1]{Equation~\ref{#1}}
\newcommand{\eqns}[1]{Equations~\ref{#1}}
\newcommand{\fig}[1]{Figure~\ref{#1}}
\newcommand{\figs}[1]{Figures~\ref{#1}}
\newcommand{\tab}[1]{Table~\ref{#1}}
\newcommand{\tabs}[1]{Tables~\ref{#1}}
\newcommand{\x}{\mathbf{x}}
\newcommand{\y}{\mathbf{y}}
\newcommand{\fid}{Fr\'echet Inception Distance\xspace}
\newcommand{\lblfig}[1]{\label{fig:#1}}
\newcommand{\lblsec}[1]{\label{sec:#1}}
\newcommand{\lbleq}[1]{\label{eq:#1}}
\newcommand{\lbltbl}[1]{\label{tbl:#1}}
\newcommand{\lblalg}[1]{\label{alg:#1}}
\newcommand{\lblline}[1]{\label{line:#1}}

\newcommand{\ignorethis}[1]{}
\newcommand{\norm}[1]{\lVert#1\rVert_1}
\newcommand{\fcseven}{$\mbox{fc}_7$}

\newsavebox\CBox
\def\textBF#1{\sbox\CBox{#1}\resizebox{\wd\CBox}{\ht\CBox}{\textbf{#1}}}
\renewcommand*{\thefootnote}{\fnsymbol{footnote}}

\def\naive{na\"{\i}ve\xspace}
\def\Naive{Na\"{\i}ve\xspace}

\makeatletter
\DeclareRobustCommand\onedot{\futurelet\@let@token\@onedot}
\def\@onedot{\ifx\@let@token.\else.\null\fi\xspace}

\def\iid{\emph{i.i.d}\onedot}
\def\eg{\emph{e.g}\onedot} \def\Eg{\emph{E.g}\onedot}
\def\ie{\emph{i.e}\onedot} \def\Ie{\emph{I.e}\onedot}
\def\cf{\emph{c.f}\onedot} \def\Cf{\emph{C.f}\onedot}
\def\etc{\emph{etc}\onedot} \def\vs{\emph{vs}\onedot}
\def\wrt{w.r.t\onedot} \def\dof{d.o.f\onedot}
\def\etal{\emph{et al}\onedot}
\def\vs{\textbf{\emph{vs}\onedot}}
\makeatother

\definecolor{citecolor}{RGB}{34,139,34}
\definecolor{mydarkblue}{rgb}{0,0.08,1}
\definecolor{mydarkgreen}{rgb}{0.02,0.6,0.02}
\definecolor{mydarkred}{rgb}{0.8,0.02,0.02}
\definecolor{mydarkorange}{rgb}{0.40,0.2,0.02}
\definecolor{mypurple}{RGB}{111,0,255}
\definecolor{myred}{rgb}{1.0,0.0,0.0}
\definecolor{mygold}{rgb}{0.75,0.6,0.12}
\definecolor{mydarkgray}{rgb}{0.66, 0.66, 0.66}

\newcommand{\yihan}[1]{{\color{purple}{[}Yihan: #1{]}}}

\newcommand\scalemath[2]{\scalebox{#1}{\mbox{\ensuremath{\displaystyle #2}}}}
\newcommand{\supplement}[1]{\color{blue}{#1}}
\newcommand{\myparagraph}[1]{\paragraph{#1}}

\def\multirowcenter{-0.5\dimexpr \aboverulesep + \belowrulesep + \cmidrulewidth}
\renewcommand{\thefootnote}{\number\value{footnote}}

\newcommand\blfootnotetext[1]{%
  \begingroup
  \renewcommand\thefootnote{}\footnotetext{#1}%
  \endgroup
}